\documentclass[letterpaper]{article} 
\usepackage{aaai2026}  
\usepackage{times}  
\usepackage{helvet}  
\usepackage{courier}  
\usepackage[hyphens]{url}  
\usepackage{graphicx} 
\usepackage{natbib}  
\usepackage{caption} 
\usepackage{algorithm}
\usepackage{algorithmic}
\usepackage{epsfig}
\usepackage{amsmath}
\usepackage{amssymb}
\usepackage{bm}
\usepackage[capitalise]{cleveref}
\usepackage{booktabs}
\usepackage{makecell}
\usepackage{multirow}
\usepackage{tabularx}
\usepackage{tikz}
\usepackage{newfloat}
\usepackage{listings}
\DeclareCaptionStyle{ruled}{labelfont=normalfont,labelsep=colon,strut=off} 
\floatstyle{ruled}
\newfloat{listing}{tb}{lst}{}
\floatname{listing}{Listing}
\newcommand{\method}{\text{LiDAR-GS++}\xspace}\usepackage{xspace}

\title{\method: Improving LiDAR Gaussian Reconstruction via Diffusion Priors}
\author{
    Qifeng Chen\textsuperscript{\rm 1}, 
    Jiarun Liu\textsuperscript{\rm 1}, 
    Rengan Xie\textsuperscript{\rm 2}, 
    Tao Tang\textsuperscript{\rm 3}, 
    Sicong Du\textsuperscript{\rm 1}, 
    Yiru Zhao\textsuperscript{\rm 1}, \\ 
    Yuchi Huo\textsuperscript{\rm 2}, 
    Sheng Yang\textsuperscript{\rm 1}\thanks{Corresponding author}
}
\affiliations{
    \textsuperscript{\rm 1}Unmanned Vehicle Dept., CaiNiao Inc., Alibaba Group\\
    \textsuperscript{\rm 2}State Key Laboratory of CAD$\&$CG Zhejiang University\\
    \textsuperscript{\rm 3}Sun Yat-sen University\\
    \{cqf7419, shengyang93fs\}@gmail.com
}

\usepackage{bibentry}

\begin{document}

\maketitle

\begin{abstract}
Recent GS-based rendering has made significant progress for LiDAR, surpassing Neural Radiance Fields (NeRF) in both quality and speed. However, these methods exhibit artifacts in extrapolated novel view synthesis due to the incomplete reconstruction from single traversal scans. To address this limitation, we present \method, a LiDAR Gaussian Splatting reconstruction method enhanced by \textit{diffusion priors} for real-time and high-fidelity re-simulation on public urban roads. 
Specifically, we introduce a controllable LiDAR generation model conditioned on coarsely extrapolated rendering to produce extra geometry-consistent scans and employ an effective distillation mechanism for expansive reconstruction.
By extending reconstruction to under-fitted regions, our approach ensures global geometric consistency for extrapolative novel views while preserving detailed scene surfaces captured by sensors.
Experiments on multiple public datasets demonstrate that \method achieves state-of-the-art performance for both interpolated and extrapolated viewpoints, surpassing existing GS and NeRF-based methods.
\end{abstract}

\begin{links}
    \link{Code}{https://github.com/CN-ADLab/LiDAR_GS_plus}
\end{links}

\section{Introduction}

Recent advances in closed-loop simulation~\cite{hu2023simulation,sss,ren2025cosmosdrivedreams} are continuously improving infrastructures for developing end-to-end autonomous driving algorithms~\cite{hu2023uniad,chen2024endtoend,fu2025orion}, where Gaussian Splatting (GS)~\cite{3dgs} has emerged as a primary representation for reconstruction-based simulators, enabling real-time rendering and ensuring physical realism. Facilitated by advancements in GS methods for camera~\cite{streetgs,drivegs} and LiDAR~\cite{gsldiar,lidargs}, developers can reconstruct urban road scenes using crowdsourced driving clips, which are typically collected to improve casually encountered corner cases, thereby meeting the requirements of arbitrary view synthesis and creating a virtual regression testing.

\begin{figure}[t]
    \centering
    \includegraphics[width=\linewidth]{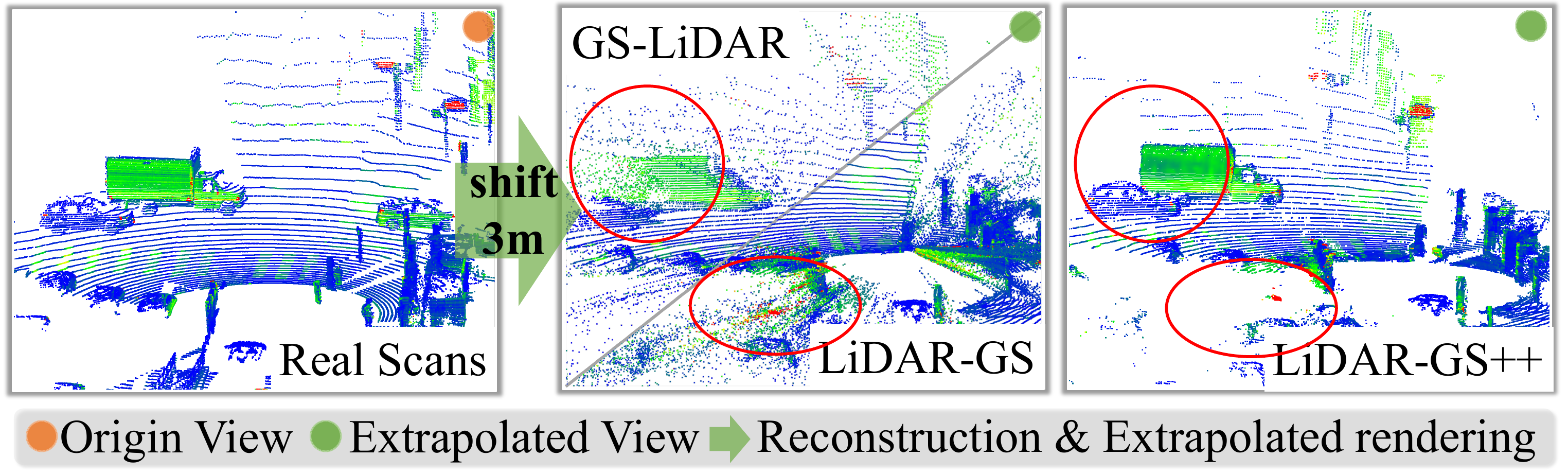}
    \vspace*{-10pt}
    \caption{LiDAR re-simulation methods, such as GS-LiDAR~\cite{gsldiar} and LiDAR-GS~\cite{lidargs}, encounter performance drops when rendering extrapolated views, e.g., lateral viewpoint shifting. In contrast, \method maintains stable and reliable performance by diffusing priors.
    }
    \label{fig:teaser}
\end{figure}%

While reconstruction-based simulators offer high realism and reusability, they are constrained by the original viewpoint distribution of the driving clip, and thus exhibit limited performance for trajectory \textit{extrapolation} when compared to \textit{interpolation}. 
While simulators for cameras~\cite{freevs,freesim,ni2025recondreamer} have made notable progress in addressing this problem, recent works on LiDAR re-simulation have not yet explicitly addressed the quality issue of extrapolation, which typically degrades simulation quality in viewpoint-changing scenarios, such as lane-changing for an obstacle avoidance. The imbalanced development of camera and LiDAR simulators decelerates the evolution of high-level (Level-4) driving agents equipped with multi-modal sensors~\cite{liao2025learning, cao2025pseudo}.

Recent advances in general controllable generative models~\cite{ldm, svd, controlnet} open up the opportunity to augment LiDAR Gaussian reconstruction with generated data, contingent upon addressing two major challenges:
(1) \textit{Authenticity}. Relying on cross-modal prompts such as texts, maps, and bounding boxes can generate promising objectives, but their scene geometry exhibit a noticeable domain gap compared to real scans that is ineffective for novel view synthesis; and (2) \textit{Consistency}. Blending generated scans with existing real scans may cause hallucinations and contradictions in specific regions, resulting in degraded harmoniousness for expansive reconstruction.

To address these challenges, we present \method and insist that LiDAR-to-LiDAR controllable generation serves as an appropriate strategy for scene expansion. Based on existing reconstruction-based simulators~\cite{lidargs,gsldiar}, we propose a novel model for conditioned LiDAR scan generation, prompted by coarsely rendered scans at extrapolated viewpoints, which improves geometry coherence and eliminates the need for auxiliary cross-modal prompts. Furthermore, given such generated scans for performing expansive reconstruction, we noticed that fully blending generated scans with existing real scans introduces confusion in regions which were already well-fitted. Thus, we introduce a depth distortion-aware distillation strategy accordingly to integrate these scans into the extrapolated reconstruction harmoniously. 

According to our experiments on multiple public datasets, \method has achieve state-of-the-art re-simulation quality on both \textit{extrapolation} and \textit{interpolation}. In conclusion, \method makes the following key contributions:

\begin{itemize}
\item We propose a LiDAR reconstruction method that expansively reconstructs neural 2DGS fields via diffusion priors, achieving state-of-the-art quality compared to previous GS-based and NeRF-based re-simulation methods in both extrapolation and interpolation.

\item For generating extrapolated LiDAR scans as extra supervision, we propose a controllable LiDAR generation model guided by coarsely extrapolated GS rendering. 

\item For expansive reconstruction, we propose a depth distortion aware distillation strategy to effectively integrate generated scans and harmoniously avoid hallucinations.

\end{itemize}

\section{Related Work}

\paragraph{Render-based LiDAR Reconstruction.}
Recently, the rapid advancement of NeRF has driven the development of a series of LiDAR re-simulation methods. NFL~\cite{Huang2023nfl} and LiDAR-NeRF~\cite{tao2023lidarnerf} pioneerly introduced differentiable frameworks that model LiDAR beams as rays, rendering depth and intensity analogous to RGB images. DyNFL~\cite{Wu2023dynfl} and LiDAR4D~\cite{zheng2024lidar4d} address dynamic reconstruction by incorporating temporal conditioning and instance-level dynamic decomposition, respectively. Mars~\cite{wu2023mars} integrates this capability into a comprehensive simulation system.

Due to the limited rendering efficiency of NeRF, several concurrent previous works have proposed to refactor the scene representation into GS:
GS-LiDAR~\cite{gsldiar} extends LiDAR4D to 2DGS representations, significantly improving inference speed. LiDAR-GS~\cite{lidargs} proposes a 3DGS-based differentiable laser beam splatting and considers the effect of ray direction and distance on rendering.
LiDAR-RT~\cite{lidarrt} extends the Boundary Volume Hierarchy (BVH) of GS-RT~\cite{gsrt} to LiDAR and implements differentiable ray tracing using the OptiX framework~\cite{parker2010optix}.

Despite these advances, existing methods have not explicitly addressed the challenge of extrapolation. To this end, we are the first to integrate diffusion priors to expand lidar Gaussian reconstruction. Our approach builds upon the efficient Gaussian attribute prediction of LiDAR-GS~\cite{lidargs}, while substituting its 3DGS representation with neural 2DGS for enhanced geometric fidelity. Comparative experiments demonstrate that this modification leads to more effective and efficient scene reconstruction.

\paragraph{Diffusion-based LiDAR Generation.}
General diffusion models~\cite{ddpm,ldm,xcube} have been demonstrated to be applicable to LiDAR generation tasks by leveraging structured representations such as 2D range view images or 3D voxel grids.
Early works, including LiDARGen~\cite{lidargen} and R2DM~\cite{r2dm}, adopt the DDPM framework for unconditional LiDAR generation.
Building on latent diffusion models (LDM)~\cite{ldm}, methods such as RangeLDM~\cite{rangeldm}, LiDM~\cite{lidm}, and Copilot4D~\cite{copilot4d} further improve generative quality by compressing LiDAR into a latent space using VAE~\cite{kingma2013vae} or VQ-VAE~\cite{van2017vqvae}, enabling more efficient training and generation.

Meanwhile, the evolution of conditional controllable generation modules~\cite{ldm, controlnet, peng2024controlnext} has further catalyzed the development for controllable LiDAR generation.
LidarDM~\cite{lidardm} employs traffic layouts to generate actors and scene. TYP~\cite{transfer} generates LiDAR Bird's Eye View(BEV) using bounding boxes and ego view. Text2LiDAR~\cite{text2lidar} pairs LiDAR scans with textual descriptions, pioneering the task of text-to-LiDAR. OLiDM~\cite{yan2025olidm} enhances object-aware LiDAR synthesis with category- and position-level prompts. UniScene~\cite{li2025uniscene} leverages an occupancy-based intermediate representation to enable consistent generation of LiDAR and camera.

While these works successfully leverage cross-modal prompts to generate desired objectives, the quality and consistency of the results against a specific driving scan remain insufficient for launching an expansive reconstruction. Given prompts of the same modality, our LiDAR-to-LiDAR controllable generation applies appropriate diffusion and denoising conditioned on the incompletely reconstructed extrapolated rendering.
Consequently, combining reconstruction and diffusion brings the opportunity to progressively enlarge the accessible LiDAR novel view synthesis range based on limited initial real scans.

\section{Method}
\begin{figure*}[ht] 
\centering
\includegraphics[width=0.90\linewidth]{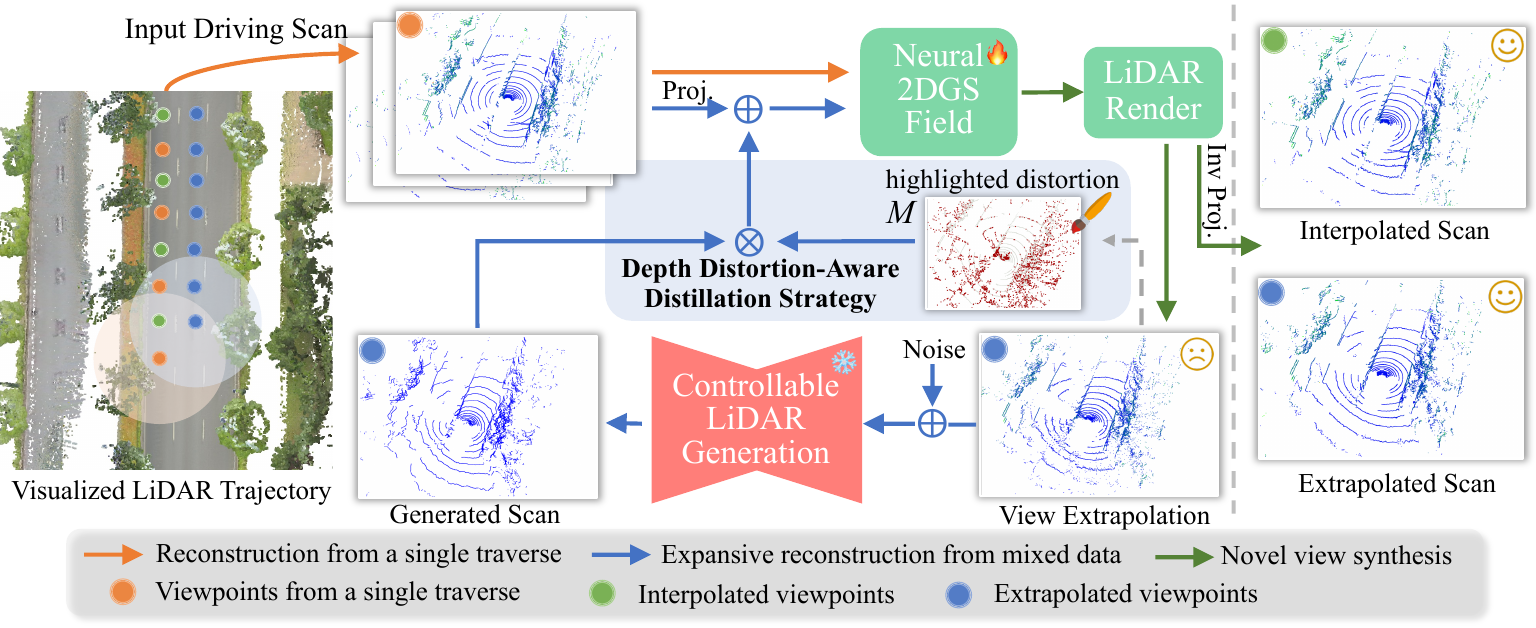}
\caption{The workflow of \method. Given an input driving scan, we first project the point cloud to the range view and reconstruct the scene using a neural 2DGS field from a single traverse. Then, we feed coarsely extrapolated rendering to a pre-trained LiDAR diffusion model to produce geometrically consistent extrapolated scans as extra supervision signals, and utilize a depth distortion-aware strategy to distill the GS representation for expansive reconstruction, where the distorted areas of the extrapolated view are highlighted in red. During inference, rendering from reconstructed Gaussians enables real-time, high-quality synthesis for both extrapolated and interpolated viewpoints. Subsequently, the range view can be converted into a point cloud format via inverse projection.}
\label{fig:workflow} 
\end{figure*}

As illustrated in \cref{fig:workflow}, we first project the point cloud to the range view and reconstruct the scene using neural 2DGS field from a single-pass driving clip. Then, we render a set of extrapolated degraded LiDAR scans by manually shifting the viewpoints, which are used as a condition for a pre-trained controllable LiDAR generative model to produce geometrically consistent additional supervision for expansive reconstruction. During expansive reconstruction, as generated and real scans are fused, we employ a Depth Distortion-Aware Distillation (DDAD) strategy on generative data to specifically correct under-fitted regions and mitigate negative effects on convergence areas during reconstruction.

\subsection{Gaussian Splatting Scene Representation} 

\paragraph{Range view Projection.} Following standard practices in LiDAR research for structurally representing distributions, we convert LiDAR data to the range view, which comprises three channels: intensity $\bar{\rho}$, depth $\bar{d}$ and ray-drop $\bar{r}$~\cite{milioto2019rangenet++, kong2023rethinking, lidargs}, where ray-drop $\bar{r} \in \{0,1\}$ indicate whether a beam has a return value. The projection process is as follows:

For each point \((x, y, z)\) in the LiDAR coordinate, its corresponding pixel location $(h, w)$ on the $\mathrm{H}\times\mathrm{W}$ range view image, as:
\begin{equation}
\begin{pmatrix}h\\w\end{pmatrix}=\begin{pmatrix}\left(1-f \left(\phi, f_b \right) \right)\\0.5 \cdot \left(1-\theta / \pi\right)\end{pmatrix} \cdot \begin{pmatrix} \mathrm{H} \\ \mathrm{W} \end{pmatrix},
\label{eq_lidar_rangeview_forward}
\end{equation}
where $\phi = \arcsin(z,\sqrt{x^2+y^2})$ and $\theta = \arctan(y,x)$ denote the vertical and horizontal angles, respectively.
The function $f \left(\phi, f_b \right)$ retrieves the position ratio closest to the vertical angle $\phi$ from a pre-defined list of vertical angles $f_b$ corresponding to each laser beam, as defined in the intrinsic parameters of the LiDAR.
Inversely, re-visualizing the point cloud can be done simply by performing inverse projection: 
\begin{equation}
(x,y,z)=d \cdot \mathbf{d} \triangleq d \cdot (\cos\theta\cos\phi,\sin\theta\cos\phi, \sin\phi),
\label{rangeview_inverse_transformation}
\end{equation}
where $d$ stands for the distance-of-flight, and the ray-direction $\mathbf{d}$ is determined by $(\phi, \theta)$. 

\begin{figure}[ht]
    \centering
    \includegraphics[width=0.9\linewidth]{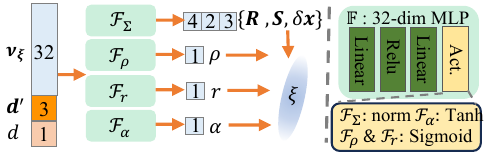}
    \caption{Design and workflow of the Neural 2DGS Field. The numbers on the vector represent the dimensions.}
    \label{fig:neuralgs}
\end{figure}

\paragraph{Neural 2DGS Field.} 
In light of the inherent distance- and direction-dependent attenuation of LiDAR returns~\cite{lidarfog,lidarsnow,lidargs} and the recent advances in geometric fidelity afforded by 2DGS~\cite{2dgs}, we propose a neural 2D Gaussian field for LiDAR scene modeling. It is composed of a set of 2D Gaussians $\xi = \{\mathbf{x},\mathbf{R}, \mathbf{S}, \rho, r, \alpha, \mathbf{v}_\xi \}$, where center position $\mathbf{x}$, rotation quaternion $\mathbf{R}$, scale $\mathbf{S}$, intensity $\rho$, ray-drop probability $r$, opacity $\alpha$ and feature token $\mathbf{v}_\xi$ (32-dims initialized to zero), respectively. 
Attributes of each Gaussian are predicted using four lightweight Multi-Layer Perceptrons (MLPs), receiving optimizable zero tokens $\mathbf{v}_\xi$, the local ray direction $\mathbf{d}'$ and distance-of-flight $d$ as input (see \cref{fig:neuralgs}). This conditioning mechanism enables the network to predict direction- and distance-dependent attributes of 2D Gaussian. Additionally the Gaussian positions $\mathbf{x}$ are refined by offsets, i.e. $\mathbf{x}\leftarrow\mathbf{x}+\delta \mathbf{x}$.

\paragraph{LiDAR Rendering.} Given the free viewpoint and orientation for rendering, we follow the 2DGS~\cite{2dgs} to compute the ray-splat intersections, and use range view rasterization to render the LiDAR's depth, intensity, and ray-drop. Specifically, at first, each 2D Gaussian $\xi$ in the world coordinate is parameterized in the local tangent plane: 
\begin{equation}
\mathbf{P}(u,v)=\mathbf{W}\mathbf{H}(u,v,1)^\top, \ \mathbf{H} \ \triangleq
\begingroup
\setlength\arraycolsep{2pt}
\begin{bmatrix}
s_u\boldsymbol{t}_u & s_v\boldsymbol{t}_v & \mathbf{x} \\
0 & 0 & 1
\end{bmatrix} \in \mathbb{R}^{4\times3},
\endgroup
\end{equation}
where $\boldsymbol{t}_u , \boldsymbol{t}_v \in \mathbb{R}^{1\times3}$ are two rotation components decomposed from rotation quaternion $\mathbf{R}$, $s_u$ and $s_v$ from $\mathbf{S}\triangleq \mathrm{diag}(s_u,s_v)$ are their corresponding scaling components, and $\mathbf{W} \in \mathbb{R}^{4\times4}$ is the view projection matrix. $\mathbf{P}(u,v)$ represents the homogeneous coordinates of a point on the local tangent plane in the world coordinate, while $(u,v)$ denotes the relative position with respect to the 2D Gaussian center in Gaussian's $uv$ coordinate.

Then, we parameterize the ray of each pixel as the intersection of two planes, $\boldsymbol{h}_u$ and $\boldsymbol{h}_v$, and the ray-splat intersection in Gaussian's local tangent plane can be determined by solving:
\begin{equation}
\begin{aligned}
&[\boldsymbol{h}_u,\boldsymbol{h}_v]^\top  \cdot \mathbf{P}(u,v)=0,\\
&\boldsymbol{h}_u=(\sin\phi,-\cos\phi,0,0)^\top,  \\
\boldsymbol{h}_v=&(\sin\theta\cos\phi, \sin\theta\sin\phi,-\cos\theta,0)^\top.
\end{aligned}
\end{equation}


Once the intersection $(u,v)$ is determined, its probability distribution value can be calculated using the standard Gaussian function
$\boldsymbol{G}_{\xi}(u,v) = \exp(-(u^2+v^2)/2)$.

During the range view rasterization process, we record the Gaussian intersections that each pixel's ray passes through, and use volume rendering integration to determine the values of the range view:
\begin{equation}
[\bar{\rho}, \bar{d}, \bar{r}] \gets \sum_{i \in \mathbf{N}} [\rho_{\xi_i}, d_{\xi_i}, r_{\xi_i}] \alpha_{\xi_i} \boldsymbol{G}_{\xi_i} \prod_{j=1}^{i-1}(1-\alpha_{\xi_j} \boldsymbol{G}_{\xi_j}),
\label{eq:rendering}
\end{equation}
where $\bar{\rho}$, $\bar{d}$, and $\bar{r}$ is the range view rendering of intensity, depth, and ray-drop, respectively.
$\mathbf{N}$ stands for the number of intersection on the ray. $\boldsymbol{G}_{\xi_i}$ denotes probability distribution value of intersection on $i$-th Gaussian.

\textbf{Optimization.} 
We optimize the neural 2DGS field, including feature token $\mathbf{v}_\xi$ of each Gaussian and parameters of four MLPs, by calculating loss between the input and rendered images:
\begin{equation}
\begin{aligned}
\mathcal{L} &=\mathcal{L}_{d}+\mathcal{L}_{\rho}+\mathcal{L}_{r}+\mathcal{L}_{\mathbf{S}}, \\
\mathcal{L}_{\rho}&=(1-\lambda_\rho) \cdot \mathcal{L}_1+\lambda_\rho \cdot \mathcal{L}_\text{D-SSIM}, \\
\end{aligned}
\label{eq:optimization}
\end{equation}
where the depth loss $\mathcal{L}_{d}$ follows $\mathcal{L}_1$ loss, and we set $\lambda_\rho = 0.2$ for the intensity loss $\mathcal{L}_{\rho}$. The ray-drop probability loss $\mathcal{L}_{r}$ follows $\mathcal{L}_2$ loss, and we use a regularization loss \cite{scaffoldgs} $\mathcal{L}_{\mathbf{S}}=\frac{1}{N} \sum_{\xi}^{N} \mathrm{Prod}( \mathbf{S}_{\xi})$ to inhibit the enlargement of scale, where $N$ is the total number of Gaussians.

\subsection{Controllable LiDAR Generation} 
In this section, we introduce a controllable LiDAR-to-LiDAR generation model designed to generate geometrically authentic extrapolated scans, serving as supplementary supervision for expansive reconstruction. A primary challenge in generative novel view synthesis lies in ensuring coherence between generation and existing knowledge. 
Existing LiDAR generation models that use cross-modal prompt, such as semantic maps and bounding boxes~\cite{lidm, transfer}, are inadequate for generating geometrically consistent scans of complex outdoor scenes, because these prompts are too sparse compared to LiDAR scans. To address this issue, we propose utilizing coarsely extrapolated LiDAR range views -- rendered from neural 2D Gaussian representations acquired during the initial stage of single-pass reconstruction -- as conditional priors for producing LiDAR scans. This approach not only establishes a robust global geometric foundation but also unifies the modalities of the condition and output, both represented as LiDAR range views.

\begin{figure}[t]
    \centering
    \includegraphics[width=\linewidth]{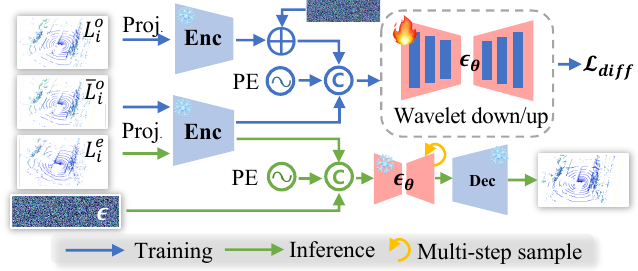}
    \caption{Workflow of the Controllable LiDAR Generation.}
    \label{fig:lidm}
\end{figure}

\paragraph{Constructing Training Pair.} 
Due to the absence of extrapolated ground truth in single-pass driving clips, direct supervision of our controllable LiDAR generation model on extrapolated viewpoints is not feasible. Hence, we simulate low-quality GS representations to construct training pairs.
Specifically, given neural 2D Gaussians reconstructed from a single-pass driving clip consisting of $F$ frames of LiDAR scans $\{L^o_i\}_{i=1}^{F}$,
we construct low-quality LiDAR rendering $\{\bar{L}^o_i\}_{i=1}^{F}$ as conditional data at the reconstructed viewpoint by introducing perturbations with a variance of $\sigma=0.2$ to the inputs of the neural 2D Gaussian field and randomly dropout Gaussian primitives at a ratio of $\tau=0.1$. 
In total, approximately 27k training pairs are constructed from Waymo Open dataset~\cite{waymo} and the Para-Lane dataset~\cite{ni2025paralane}, taking 20h on 4$\times$RTX3090.

\paragraph{Training Diffusion Model.}
We choose the Latent LiDAR Diffusion Model (LiDM)~\cite{lidm} as our base model, and introduce two additional lightweight modules to improve generation details, including adding fourier Position Encoding (PE) at input \cite{rangeldm} and integrating wavelet transforms into the upsampling and downsampling layer \cite{dpmms}. As shown in \cref{fig:lidm}, the LiDAR's input and condition are respectively encoded into latent space as $z^L$ and $c^{\bar{L}}$ with a pre-trained VAE encoder. After diffusing the input $z^L$ to $z^L_t = \sqrt{\bar{\alpha}_t} z^L_0 + \sqrt{1-\bar{\alpha}_t} \epsilon$ with noise $\epsilon\sim\mathcal{N}(\mathbf{0},\boldsymbol{I})$ and a noise schedule $\bar{\alpha}$ at diffusion time step $t$, we concatenate them as the input of denoised model $\epsilon_\theta$ and optimize it by the loss function:
\begin{equation}
\mathcal{L}_{diff}=\mathbb{E}_{z^L_0,\epsilon,c^{\bar{L}},t}\left[\|\epsilon -\epsilon_\theta(z^L_t,t,c^{\bar{L}})\|_2^2\right].
\end{equation}

This design enables recovery of realistic LiDAR from low-quality cues and generalizes to produce scans from extrapolated coarse rendering.
\paragraph{Generated LiDAR Scan on View Extrapolation.} Given a set of 2D Gaussians $\xi$ reconstructed from a single-pass driving clip, we render the low-quality LiDAR scans $\{L^e_i\}_{i=1}^{K}$ across lanes by extrapolating the viewpoint. These scans serve as conditional guidance for the frozen LiDAR generation model to produce geometrically consistent extrapolated scans, which provide additional supervision to refine the neural 2D Gaussians $\xi$.

\subsection{Depth Distortion-Aware Distillation Strategy} 
\begin{figure}[ht]
    \centering
    \includegraphics[width=\linewidth]{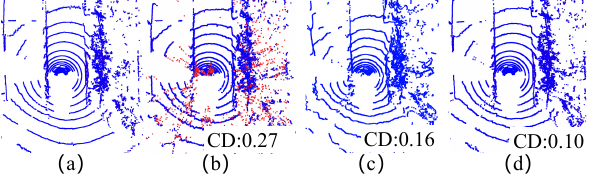}
    \caption{
    Qualitative and quantitative comparisons of our coarse-to-fine cross-lane novel view synthesis on the Para-Lane dataset. (a) Real LiDAR scan as a reference from another-pass; (b) Rendered without diffusion priors, and the red point indicates the distorted area $M$; (c) Generative result of controllable LiDAR diffusion model conditioned by coarsely extrapolated rendering; (d) Rendering after expansive reconstruction with diffusion priors.}
    \label{fig:ddas}
\end{figure}

Even though the generative model can eliminate most of the incorrect LiDAR rendering, it still introduces hallucinations in the details (See \cref{tab:ablation} for ablation experiments of expansive reconstruction without DDAD). As shown in the \cref{fig:ddas} (c), we use the Chamfer Distance (CD) to quantify the difference between the generative results and the reference real scans on the Para-Lane dataset, revealing that there are still detail discrepancies between them. In fact, the \cref{fig:ddas} (b) has demonstrated that neural 2D Gaussians have preliminary ability to render global geometry in extrapolated viewpoints, with artifacts and rendering collapses occur in under-fitted regions due to insufficient extrapolated supervision.
As shown in \cref{tab:ablation}, if we fully inject generated data, the upper bound for novel views is limited by the quality of the generative data (as \cref{fig:ddas} (c)), and negative effects may also be introduced to already well-fitted regions.

Therefore, we propose a depth distortion-aware distillation strategy specifically designed to identify and correct under-fitted regions in extrapolated views, thereby minimizing the adverse impact of the LiDAR generative model for GS expansive reconstruction.
Specifically, during the rendering phase, we record the median depth $d_m$ for each ray, approximated by the depth at the Gaussian intersection where the transmittance $T$ is closest to 0.5~\cite{mipnerf360, 2dgs}. Following~\cref{eq:rendering}, transmittance $T$ is defined as follows:
\begin{equation}
T=\prod_{j=1}^{i-1}(1-\alpha_{\xi_j} \boldsymbol{G}_{\xi_j}).
\label{eq:transmittance}
\end{equation}

It is well-established that when transmittance approaches 0, the ray is considered to have passed through the physical surface, and rendering halts, outputting the rendered depth $\bar{d}$. Generally, Gaussian distributions that converge to the real depth contribute significantly to the rendering when their cumulated opacity close to 1. In these cases, as indicated by \cref{eq:rendering}, the discrepancy between the median depth $d_m$ and rendered depth $\bar{d}$ is minimal. Conversely, a large difference suggests that the attributes of the Gaussians involved in rendering remain under-fitted. Therefore, we define areas with significant depth discrepancies as distortion regions:
\begin{equation}
    M = \{| \bar{d}_m-\bar{d} |>\delta\},
\label{eq:M}
\end{equation}
where $\delta=\mathrm{median}\{\mathrm{max}(s_u,s_v) \}$ is defined as the median value of the scale coefficients from the longest axis of all Gaussian trained from single traverse.

In \cref{fig:ddas} (b), this definition effectively labels ambiguous rendered point clouds as distorted regions $M$, allowing us to further fine-tune the Gaussian in a targeted manner. Specifically, when mixing the generated extrapolated scans as additional supervision to further expansive reconstruction(1:1 with real), the loss function is following $\mathcal{L}_e=M\cdot\mathcal{L}$.

\section{Experiment}

\begin{table*}[ht]
\centering
\scriptsize
\setlength{\tabcolsep}{2.4pt}
\begin{tabularx}{0.98\linewidth}{c|cccccccc|cccccccc|cc}
\toprule
  & \multicolumn{8}{c|}{Para-Lane} & \multicolumn{8}{c|}{Waymo} & \multicolumn{2}{c}{Cost} \\ 
 \cmidrule{2-19} 
\multicolumn{1}{c|}{Method}&  \multicolumn{4}{c|}{Lane Extrapolation @3.5m}  & \multicolumn{4}{c|}{Lane Interpolation} &  \multicolumn{4}{c|}{Lane Extrapolation @3.5m}  & \multicolumn{4}{c|}{Lane Interpolation} & 
\multicolumn{1}{c}{Train} &
\multicolumn{1}{c}{Inference} \\
\cmidrule{2-19} 
& CD$\downarrow$ & F-scr$\uparrow$ & PSNR$\uparrow$  & \multicolumn{1}{c|}{SSIM$\uparrow$} & CD$\downarrow$ & F-scr$\uparrow$ & PSNR$\uparrow$  & SSIM$\uparrow$ & FRID$\downarrow$ & FPVD$\downarrow$ & JSD$\downarrow$ & \multicolumn{1}{c|}{MMD$\downarrow$} & CD$\downarrow$ & F-scr$\uparrow$ & PSNR$\uparrow$  & SSIM$\uparrow$ & (min))$\downarrow$ & (fps) $\uparrow$ \\ \midrule
LiDAR4D & 1.518 & 0.785 & 29.464          & \multicolumn{1}{c|}{0.833}          & 0.112          & 0.921           & 31.631          & 0.866          & 48.503           & 52.651           & 0.587           & \multicolumn{1}{c|}{9.726}           & 0.061          & 0.964           & 32.434          & 0.970          & \multicolumn{1}{c}{426}                                     & 1.7                                       \\
LiDAR-RT                & 0.482          & 0.806           & 30.430          & \multicolumn{1}{c|}{0.824}          & 0.159          & 0.902           & 30.979          & 0.837          & 41.330           & 57.551           & 0.576           & \multicolumn{1}{c|}{11.738}          & 0.063          & 0.962           & 29.410          & 0.917          & \multicolumn{1}{c}{213}                                     & \textbf{20.7}                                      \\
GS-LiDAR                & 0.305          & 0.843           & 29.279          & \multicolumn{1}{c|}{0.802}          & 0.086          & 0.927           & 29.654          & 0.805          & 31.967           & 78.84            & 0.534           & \multicolumn{1}{c|}{8.263}           & 0.057          & 0.971           & 31.732          & 0.952          & \multicolumn{1}{c}{129}                                     & 10.8                                      \\
LiDAR-GS                & 0.270          & 0.865           & 30.742          & \multicolumn{1}{c|}{0.828}          & 0.090          & 0.923           & 32.103          & 0.869          & 39.095           & 34.018           & 0.606           & \multicolumn{1}{c|}{13.301}          & 0.059          & 0.968           & 32.313          & 0.973          & \textbf{18}    & 15.8                                      \\ \midrule
\textbf{\method}     & \textbf{0.102} & \textbf{0.923}  & \textbf{31.843} & \multicolumn{1}{c|}{\textbf{0.871}} & \textbf{0.079} & \textbf{0.936}  & \textbf{32.232} & \textbf{0.873} & \textbf{11.669}  & \textbf{15.134}  & \textbf{0.379}  & \multicolumn{1}{c|}{\textbf{2.618}}  & \textbf{0.052} & \textbf{0.972}  & \textbf{32.463} & \textbf{0.975} & \multicolumn{1}{c}{26}                                      & 16.2                                      \\ \bottomrule
\end{tabularx}
\caption{Average quantitative results on the Para-Lane dataset and Waymo Open dataset.The MMD metric is at the $1e-5$ scale.}
\label{table:compare_result}
\end{table*}

\begin{figure*}[!ht] 
\centering 
\includegraphics[width=0.98\linewidth]{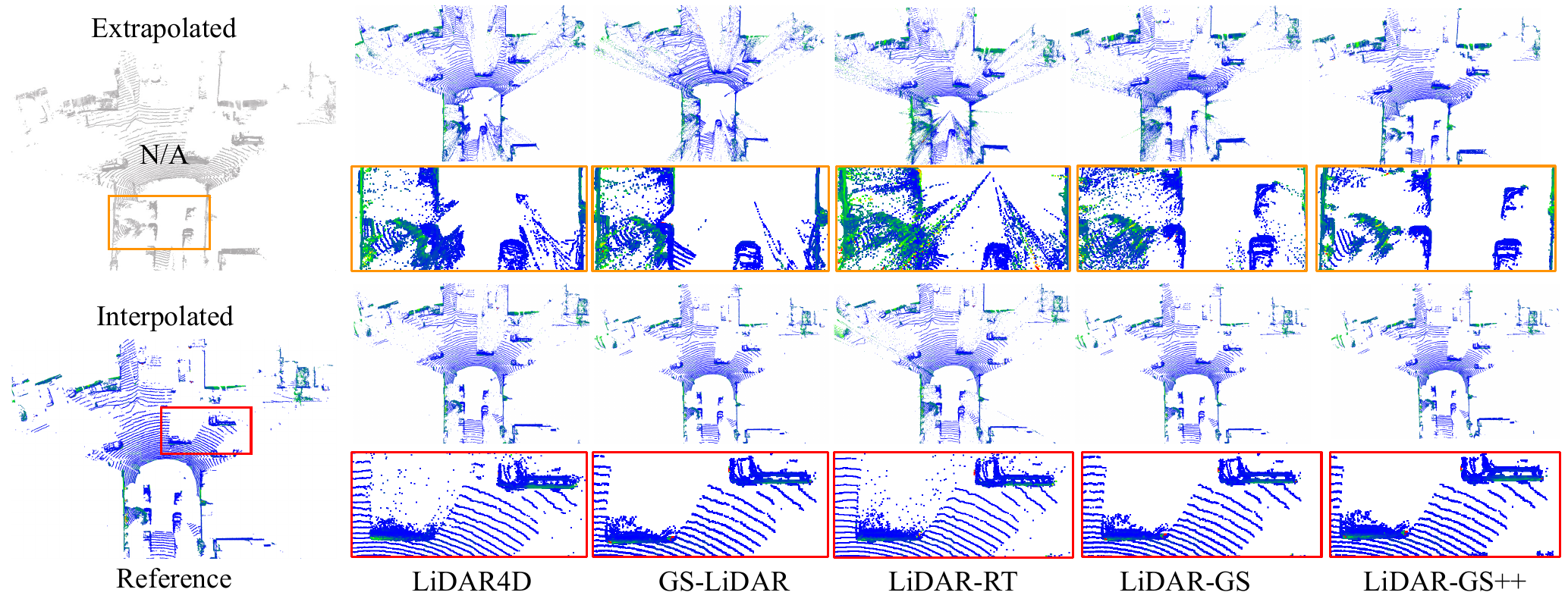}
\caption{Qualitative comparison on Waymo sequences. Points are colorized through their intensity.}
\label{fig:compare_result} 
\end{figure*}

\subsection{Experiment Setup}
\paragraph{Dataset.}
We conduct experiments on both the Para-Lane~\cite{ni2025paralane} and the Waymo Open~\cite{waymo} datasets. The Para-Lane dataset provides synchronized cross-lane LiDAR data, with a resolution of 32$\times$1800 at 10 Hz. The Waymo dataset provides LiDAR data along a single lane with a resolution of 64$\times$2650 at 10 Hz. We selected 3 clips from the Para-Lane dataset and 4 clips from the Waymo dataset to validate our method against publicly available approaches for novel view synthesis, including both viewpoint interpolation and extrapolation.

\paragraph{Baseline.}
We compare our method with LiDAR4D~\cite{zheng2024lidar4d}, GS-LiDAR~\cite{gsldiar}, LiDAR-RT~\cite{lidarrt} and LiDAR-GS~\cite{lidargs}. LiDAR4D, to our best knowledge, is currently the best NeRF-based LiDAR synthesis method. Meanwhile, the remaining methods are the latest advancements in real-time synthesis based on GS. Please refer to our \textit{supplementary material} for detailed configurations.

\paragraph{Implementation details of Gaussian representation.}
\method was trained for 7,000 iterations on RTX3090, with the first 5,000 iterations dedicated to reconstruction from single traverse and the last 2,000 iterations to fine reconstruction from mixed generation and real data. We randomly initialized 500,000 GS anchors from trained scan input and adopted the same growth and split strategy as Scaffold-GS~\cite{scaffoldgs} and the same accumulated gradient strategy as AbsGS~\cite{ye2024absgs}. During densification, we enable it starting from step 500 with the split gradient threshold set to 0.002. We use the Adam optimizer and set the learning rate for each optimizable parameter $\{\mathcal{F}_{{\Sigma}}$, $\mathcal{F}_\rho$, $\mathcal{F}_{r}$, $\mathcal{F}_{\alpha}, \mathbf{v}_\xi \}$ to $\{1,4,4,2,5\} \times 1e^{-3}$, respectively. For dynamic objects, we perform instance decomposition based on NSG~\cite{nsg} and reconstruct them separately.

\paragraph{Implementation details of controllable LiDAR generation model.} 
The diffusion module with a frozen VAE is trained once (not per-scene) on constructed pairs for $5e^4$ iterations using 8$\times$A100 GPUs and batch size of 16 on each GPU . We use the Adam optimizer and set the learning rate to  $5e^{-5}$. During inference phase, we use the DDIM~\cite{ddim} sampler with 50 sampling steps.

\subsection{Main Results}

\paragraph{Quantitative comparison.} On the Para-Lane dataset, we sample every 10th viewpoint from the input of a single trajectory as interpolation evaluation, and use the remaining scans as training inputs. Since this dataset provides synchronized ground-truth LiDAR scans from adjacent lanes, we use viewpoints from neighboring lanes for evaluating extrapolation. Following previous work~\cite{lidargs}, we use CD and F-Score as evaluation metrics for point clouds, and employ PSNR and SSIM as evaluation metrics for LiDAR intensity and ray-drop.

On the Waymo dataset, the evaluation for interpolation is conducted in the same manner as described above. The difference is that, due to the lack of extrapolated ground truth, we artificially shift the input viewpoints laterally (@3.5m) as extrapolated viewpoints, and adopt the generative evaluation metrics as LiDM~\cite{lidm}, including Fréchet Range Image Distance (FRID), Fréchet Point-based Volume Distance (FPVD), Minimum Matching Distance (MMD), Jensen-Shannon Divergence (JSD). Refer to \textit{supplementary materials} for details.

As shown in \cref{table:compare_result}, \method achieves excellent performance in interpolation evaluation while significantly outperforming previous methods in extrapolation evaluation. We owe this superior performance to extrapolation supervision from the diffusion prior and the expansive reconstruction of distorted areas, which enable a better representation of the neural 2DGS field.
Meanwhile, our method achieves suboptimal computing efficiency. In terms of training speed, \method is only slower than LiDAR-GS due to additional computational overhead incurred during expansive reconstruction. Regarding inference speed, \method is second only to LiDAR-RT, which benefits from OptiX's~\cite{parker2010optix} hardware acceleration, enabling faster ray tracing on specific hardware configurations.

\begin{table}[ht]
\centering
\scriptsize
\setlength{\tabcolsep}{1.9pt}
\begin{tabularx}{\linewidth}{c|cccc|cccc}
\toprule
 \multirow{2}{*}{Method}
&  \multicolumn{4}{c|}{Lane Extrapolation @3.5m}  & \multicolumn{4}{c}{Lane Interpolation} \\ \cmidrule{2-9} 
         & CD$\downarrow$ & F-scr$\uparrow$  & PSNR$\uparrow$  & SSIM$\uparrow$  & CD$\downarrow$ & F-scr$\uparrow$  & PSNR$\uparrow$  & SSIM$\uparrow$ \\ \midrule
\makecell{GS-LiDAR \\ (w/ Diff\&DDAD)}
              & 0.116 & 0.921 & 29.530 & 0.816 & 0.087 & 0.925 & 29.654 & 0.804  \\ \midrule
Ours w/o NGF  & 0.417 & 0.825 & 29.878 & 0.809 & 0.095 & 0.922 & 30.039 &   0.816  \\
Ours w/o Diff & 0.264 & 0.869 & 30.777 & 0.840 & 0.079 & \textbf{0.937} & \textbf{32.235} & 0.873 \\
Ours w/o DDAD & 0.163 & 0.905 & 30.701 & 0.839 & 0.085 & 0.927 & 32.013 & 0.854 \\
\midrule
Ours & \textbf{0.102} & \textbf{0.923} & \textbf{31.843} & \textbf{0.871} & \textbf{0.079} & 0.936 & 32.232 & \textbf{0.873}  \\
\bottomrule
\end{tabularx}
\caption{Ablation study on Para-Lane dataset.}
\label{tab:ablation}
\end{table}

\begin{table}[ht]
\centering
\scriptsize
\setlength{\tabcolsep}{4pt}
\begin{tabularx}{0.98\linewidth}{c|cccc}
\toprule
Method & FRID$\downarrow$ & FPVD$\downarrow$  & JSD$\downarrow$ & MMD$\downarrow$($1e^{-5}$)\\ 
\midrule
CLG w/o PE & 33.62 & 15.17 & 0.51 &  13.76   \\
CLG w/o Wavelet & 29.31 & 14.88 & 0.50 &  10.37 \\
\midrule
CLG + BBox + Map Condition  & 46.74 & 109.51 & 0.66 & 17.72  \\
\midrule
CLG + Render Condition (Ours) & \textbf{28.39} & \textbf{14.46} & \textbf{0.48} &  \textbf{6.20} \\
\bottomrule
\end{tabularx}
\caption{Ablation study on Controllable LiDAR Generation.}
\label{tab:ablation_generation}
\end{table}

\paragraph{Qualitative comparison.} \cref{fig:compare_result} shows the visual comparison of LiDAR rendering. In detail, others perform poorly under extrapolated viewpoints, exhibiting numerous floater and incorrect geometry, whereas our method maintains stable and reliable performance for extrapolated re-simulation. 

\subsection{Ablation Results}

\paragraph{Ablation on components of \method.} We firstly replaced our used neural 2DGS fields (NGF) with the vanilla 2DGS~\cite{2dgs}. The results presented in \cref{tab:ablation} indicate that 
 the neural 2DGS field achieves superior reconstruction performance compared to previous Gaussian-based methods.Moreover, as shown in Table 1, our approach outperforms LiDAR-GS~\cite{lidargs} based 3DGS. 
 These results underscore the critical role of the neural 2DGS field, as a robust GS representation serves as the essential foundation for achieving high-fidelity LiDAR scene reconstruction.
 This imporovement is attributed to consideration of the impact of viewpoint and distance on LiDAR signals and the use of 2DGS primitives that are more advantageous for geometric reconstruction.

Then, to assess the necessity of incorporating diffusion priors, we conducted an experiment where extrapolated viewpoints were rendered without employing diffusion priors (w/o Diff) as additional supervision. Results in~\cref{tab:ablation} demonstrate that the introduction of extrapolated scans generated by controllable LiDAR generation models, as supplementary supervision for expansive reconstruction, can significantly enhance the quality of extrapolated view.

Finally, we compared the performance between utilizing fully generated data for expansive reconstruction and applying Depth Distortion-Aware Distillation (DDAD) strategy. As shown in~\cref{tab:ablation}, the integration of the DDAD strategy yields superior results. Because the use of generated data without any careful adjustment not only limits the upper bound of extrapolated reconstruction but also introduces adverse effects on the real training viewpoints. As illustrated in~\cref{fig:ddas}, there are still detailed discrepancies between the generated data and the real reference. Our proposed strategy selectively optimizes the under-fitted regions during expansive reconstruction, preserving the convergent geometric details from the initial single traversal reconstruction.

\paragraph{Analysis of the effectiveness of diffusion priors.} 
We integrated the diffusion prior into the previous sota, GS-LiDAR~\cite{gsldiar}, for expansive reconstruction, which achieve substantial improvements when evaluated from extrapolation perspective, as illustrated in Table 3. These results demonstrate the effectiveness and applicability of our proposed diffusion prior for LiDAR reconstruction.

\begin{figure}[t]
    \centering
    \includegraphics[width=0.98\linewidth]{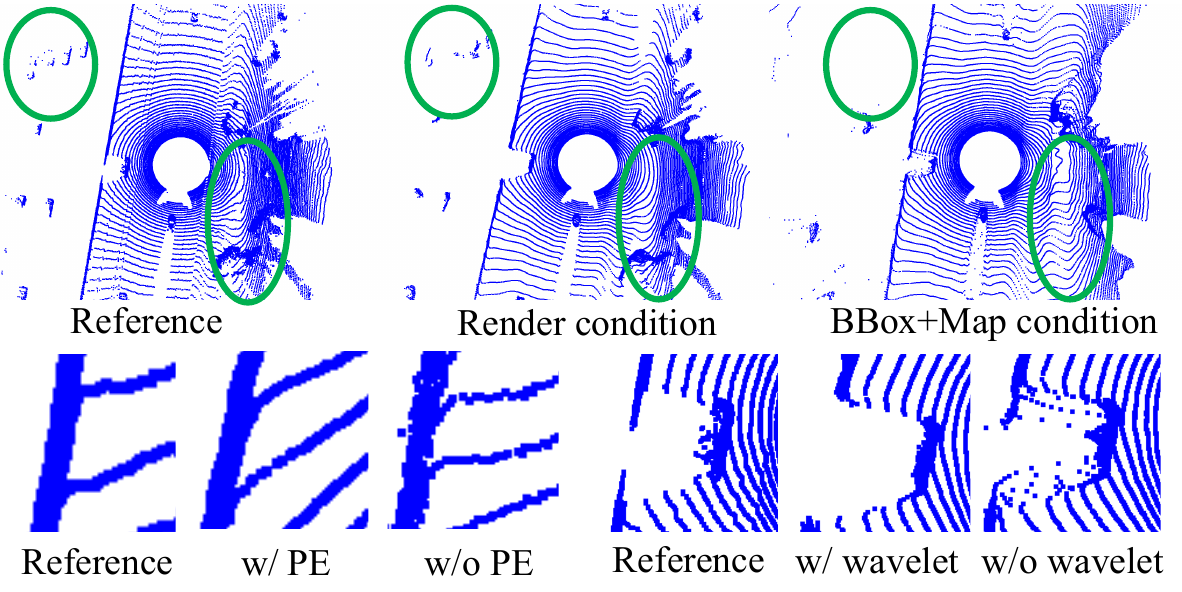}
    \caption{Qualitative visualization of CLG ablation.}
    \label{fig:CLG}
\end{figure}

\paragraph{Analysis on Controllable LiDAR Generation (CLG).}
The results of the two specific ablation experiments conducted for CLG indicate that the integration of the additional modules provides positive contributions. As shown in~\cref{fig:CLG}, incorporating wavelet transformation aids in preserving high-frequency details during downsampling and upsampling processes, and the introduction of PE facilitates the network's learning of continuous geometric encoding. Additionally, following~\cite{lidm, transfer}, we trained a cross-modal LiDAR generation model conditioned on semantic maps and bounding boxes using the same driving clip, and compared it to our single-modal LiDAR-to-LiDAR generation model. Our model demonstrated greater potential for LiDAR synthesis tasks. Moreover, the figures reveal that although the diffusion model conditioned on semantic maps and bounding boxes shows commendable generative performance for specified prompts, it often produces geometric inconsistencies in areas lacking dense prompts, such as trees.

\section{Conclusion}

This paper proposes \method, a novel LiDAR Gaussian Splatting reconstruction framework enhanced by \textit{diffusion priors}, designed to enable real-time and high-fidelity re-simulation on public urban roads. \method leverages a controllable LiDAR generation model guided by coarsely extrapolated rendering to provide consistent additional supervision for expansive reconstruction. And we further propose a depth distortion-aware distillation strategy to efficiently refine the neural 2DGS representation in under-fitted regions. Through extensive evaluations on publicly available datasets, \method demonstrates state-of-the-art performance on both interpolated and extrapolated viewpoints.

\paragraph{Limitations.} First, the reconstruction process doesn't account for non-rigid motion of dynamic objects, e.g. pedestrians. Second, LiDAR generative models lack consideration for temporal consistency. Future work could explore the use of more advanced generation models to improve fidelity and consistency in synthesized LiDAR data.

\section{Acknowledgments}
We thank the reviewers for the valuable discussions. This research was supported by the Zhejiang Provincial Natural Science Foundation of China under Grant No. LD24F030001.

\small
\bibliography{main}

\end{document}